\documentclass[10pt,twocolumn,letterpaper]{article}

\usepackage{cvpr}             
\usepackage{graphicx}
\usepackage{amsmath}
\usepackage{amssymb}
\usepackage{booktabs}
\usepackage{mathtools, cuted}
\usepackage{mathrsfs}
\usepackage{breqn}

\usepackage{array}
\usepackage{amsfonts}
\usepackage[margin=1.0in]{geometry}
\usepackage{tabularx}
\usepackage{multirow}
\usepackage{url}
\usepackage{makecell}
\newcolumntype{P}[1]{>{\centering\arraybackslash}p{#1}}
\usepackage{caption}
\usepackage{subcaption}
\usepackage[pagebackref,breaklinks,colorlinks]{hyperref}
\usepackage[capitalize]{cleveref}
\crefname{section}{Sec.}{Secs.}
\Crefname{section}{Section}{Sections}
\Crefname{table}{Table}{Tables}
\crefname{table}{Tab.}{Tabs.}

\begin{document}

\title{Enhancing ResNet Image Classification Performance by using Parameterized Hypercomplex Multiplication}

\author{Nazmul Shahadat, Anthony S.\ Maida\\
University of Louisiana at Lafayette\\
Lafayette LA 70504, USA\\
{ nazmul.ruet@gmail.com, maida@louisiana.edu}}
\maketitle

\begin{abstract}
Recently, many deep networks have introduced hypercomplex and related calculations into their architectures. 
In regard to convolutional networks for classification, these enhancements have
been applied to the convolution operations in the frontend to enhance accuracy and/or reduce the parameter requirements while maintaining accuracy.
Although these enhancements have been applied to the convolutional frontend, it has
not been studied whether adding hypercomplex calculations improves performance when applied
to the densely connected backend. 
This paper studies ResNet architectures and 
incorporates parameterized hypercomplex multiplication (PHM) into the backend of residual, quaternion, and vectormap convolutional neural networks to
assess the effect.
We show that PHM does improve classification accuracy performance
on several image datasets, including small, low-resolution CIFAR 10/100 and large 
high-resolution ImageNet and ASL,
and can achieve state-of-the-art accuracy for hypercomplex networks.
\end{abstract}

\section{Introduction}
\label{sec:intro}
\begin{figure}[ht] 
\centering
\begin{subfigure}{.45\textwidth}
  \centering
  \includegraphics[width=.9\textwidth]{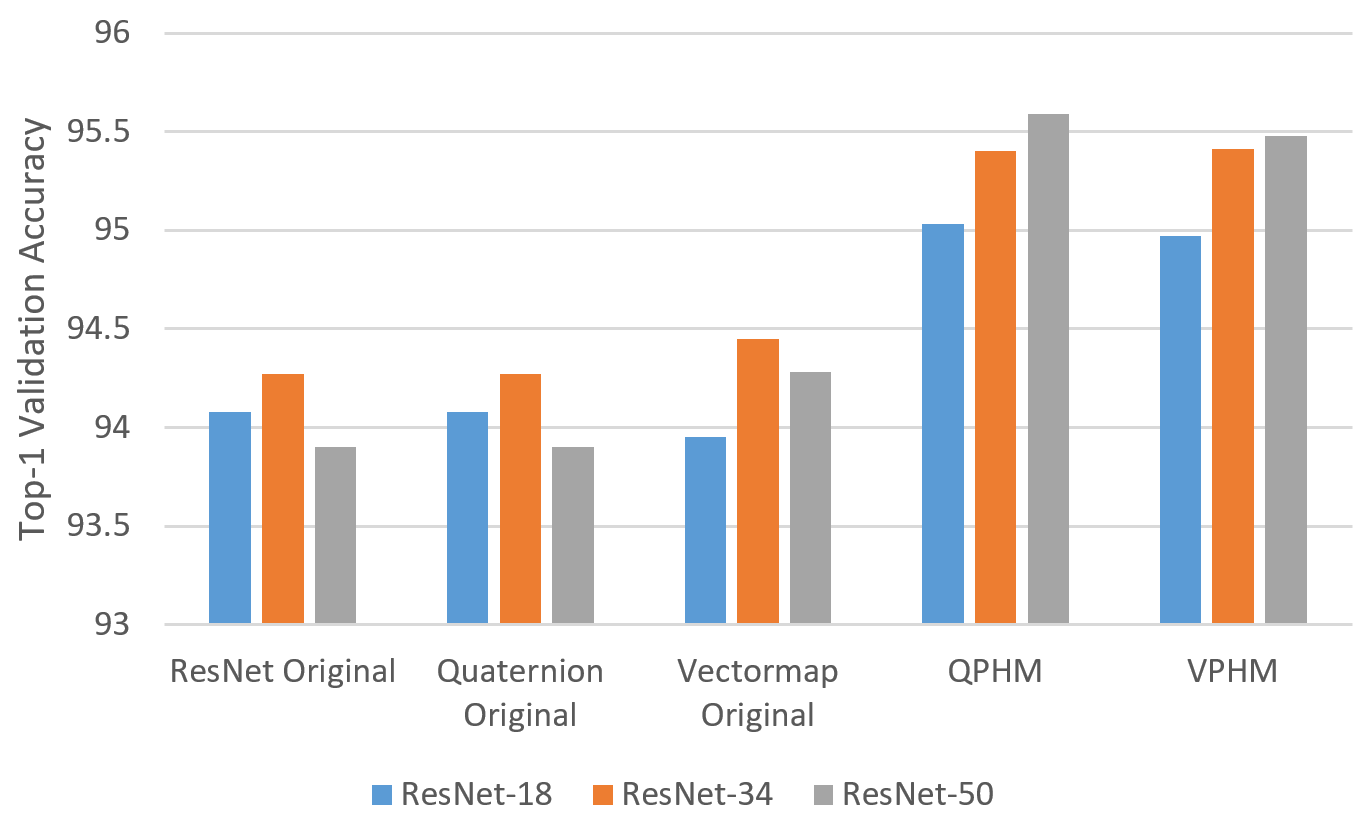}
  \captionsetup{width=0.95\textwidth}
  \captionof{figure}{Validation accuracy comparison for CIFAR-10 data.}
  \label{fig:fig1CIFAR10}
\end{subfigure} 
\begin{subfigure}{.45\textwidth}
  \centering
  \includegraphics[width=.9\textwidth]{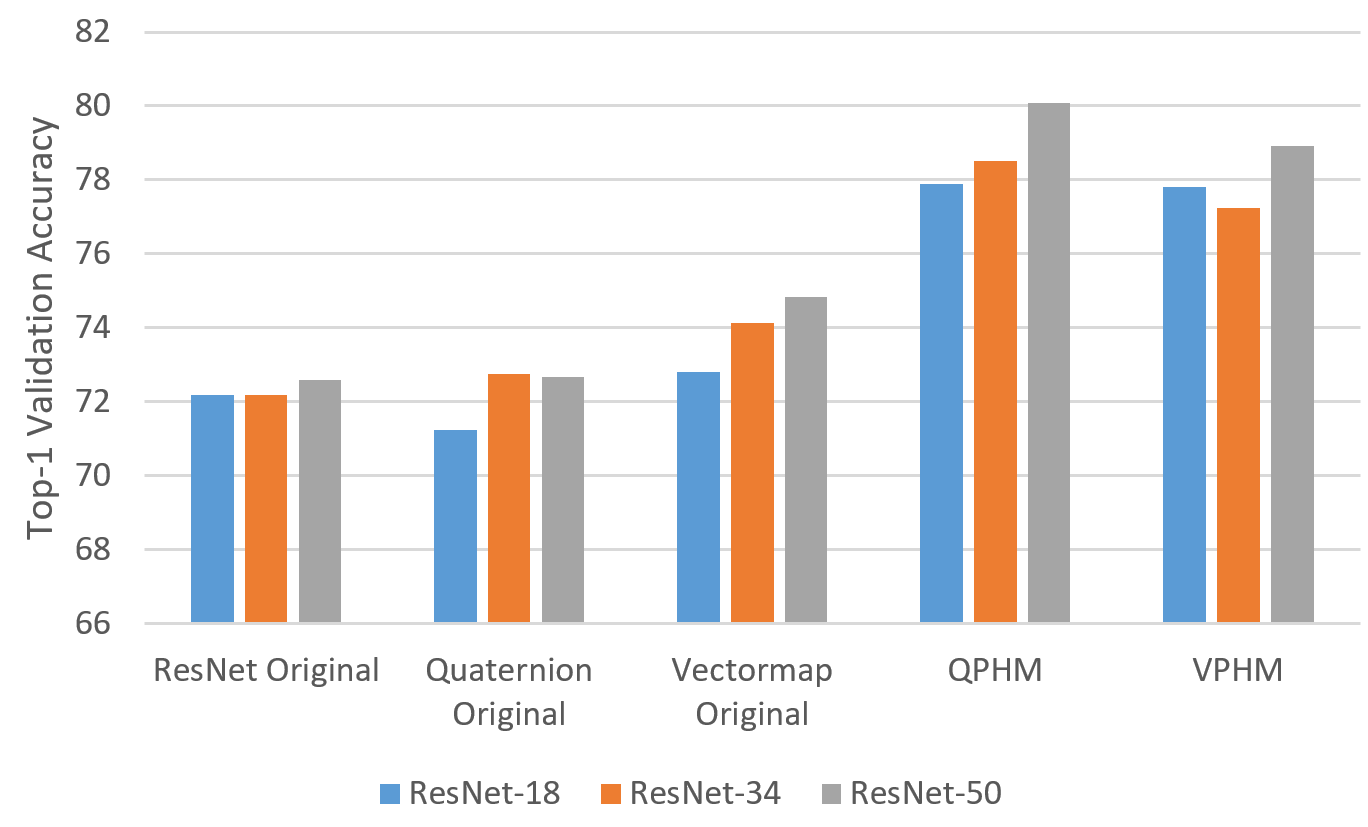}
  \captionsetup{width=0.95\textwidth}
  \captionof{figure}{Validation accuracy comparison for CIFAR-100 data.}
  \label{fig:fig2CIFAR100}
\end{subfigure}
\caption{Top-1 validation accuracy comparison among original ResNets \cite{gaudet2021removing}, original quaternion networks \cite{gaudet2021removing}, original vectormap networks \cite{gaudet2021removing}, our proposed QPHM and VPHM networks for CIFAR benchmarks}
\label{fig_Fig1_CIFAR10_and100}
\end{figure}
Convolutional neural networks (CNNs) have been widely used, with great success,
in visual classification tasks \cite{buyssens2012multiscale,javanmardi2021computer} because of their good inductive priors and intuitive design.

Most deep learning building blocks in CNNs use real-valued operations.
However, recent studies have explored the complex/hypercomplex space and showed that hypercomplex valued  networks can perform better than their real-valued counterparts due to the weight sharing mechanism 
embedded in the hypercomplex multiplication
\cite{gaudet2021removing,parcollet2019quaternion}. 
This weight sharing differs from that found in the real-valued convolution operation.
Specifically,
quaternion convolutions share weights across input channels
enabling them to discover cross-channel input relationships that support more
accurate prediction and generalization. 
The effectiveness of quaternion networks is shown in \cite{gaudet2021removing,parcollet2018quaternion,long2019quaternion,shahadat2021adding,zhu2018quaternion}.

The weight-sharing properties of the Hamiltonian product allow
the discovery of cross-channel relationships.
This is a new plausible inductive bias, namely,
that there are data correlations across convolutional input channels that enhance
discovery of effective cross-channel features. Practitioners have applied these calculations in the convolution stages of CNNs but not to the 
dense backend where real-valued operations are still used. 
The present paper
puts weight-sharing calculations in the dense backend to further improve CNN performance. To exploit this new type of weight sharing, we use a parameterized hypercomplex multiplication (PHM) \cite{zhang2021beyond} layer as a building block.
This block replaces the real-valued FC layers with hypercomplex FC layers. 
We test the hypothesis using two types of hypercomplex CNNs, namely
quaternion \cite{gaudet2018deep} CNNs and vectormap \cite{gaudet2021removing} CNNs.

Our contributions are:
\begin{itemize}
    \item Showing the effectiveness of using hypercomplex networks in the densely connected backend of a CNN.
    \item Introducing quaternion networks with PHM based dense layer (QPHM) to bring hypercomplex deep learning properties to the entire model.
    \item Introducing vectormap networks with a PHM based dense layer (VPHM) to remove hypercomplex dimensionality constraints
          from the frontend and backend.
\end{itemize}

The effectiveness of employing PHM based FC layers with hypercomplex networks is seen in Figures \ref{fig:fig1CIFAR10} and~\ref{fig:fig2CIFAR100}. We also show that these new models obtain SOTA results for hypercomplex networks in CIFAR benchmarks.
Our experiments also show SOTA results for American Sign Language (ASL) data. Moreover, our models use fewer parameters, FLOPS, and latency compared to the base model proposed by \cite{gaudet2021removing,shahadat2021adding} for classification.

\section{Background and Related Work}
\subsection{Quaternion Convolution}
\label{subsection_quatconv}

Quaternions are four dimensional vectors of the form
\begin{equation} 
Q =  r + \textit{i}x + \textit{j}y + \textit{k}z~;~r,x,y,z \in \mathbb{R} 
\end{equation}
where, $r$, $x$, $y$, and $z$ are real values and $i$, $j$, and $k$ are the imaginary values which satisfy
$i^2=j^2=k^2=ijk=-1$. Quaternion convolution is defined by convolving a quaternion filter matrix with a quaternion vector (or feature map). 
Let, $Q_F = \mathbf{R} + i\mathbf{X} + j\mathbf{Y} + k\mathbf{Z}$ be a quaternion filter matrix 
with $\mathbf{R}$, $\mathbf{X}$, $\mathbf{Y}$, and $\mathbf{Z}$ being real-valued kernels
and $Q_V = \mathbf{r} + i\mathbf{x} + j\mathbf{y} + k\mathbf{z}$ be a 
quaternion input vector with
$\mathbf{r}$, $\mathbf{x}$, $\mathbf{y}$, and $\mathbf{z}$ being real-valued vectors. 
Quaternion convolution is defined below \cite{gaudet2018deep}.
\begin{equation}
    \begin{aligned}
        Q_F \circledast Q_V =  (\mathbf{R}*\mathbf{r} - \mathbf{X}*\mathbf{x} - \mathbf{Y}*\mathbf{y} -\mathbf{Z}*\mathbf{z})\\
        + \textit{i}(\mathbf{R}*\mathbf{x} + \mathbf{X}*\mathbf{r} + \mathbf{Y}*\mathbf{z} - \mathbf{Z}*\mathbf{y})\\
        + \textit{j}(\mathbf{R}*\mathbf{y} - \mathbf{X}*\mathbf{z} + \mathbf{Y}*\mathbf{r} + \mathbf{Z}*\mathbf{x})\\ 
        + \textit{k}(\mathbf{R}*\mathbf{z} + \mathbf{X}*\mathbf{y} - \mathbf{Y}*\mathbf{x} + \mathbf{Z}*\mathbf{r})
        \label{quat:10}
    \end{aligned} 
\end{equation}

\noindent
There are 16 real-valued convolutions but only four kernels which are reused.
This is how the weight sharing occurs.
\cite{parcollet2019quaternion} first described the weight sharing in the Hamilton product.

\subsection{Vectormap Convolution}
\label{subsection_vectconv}

\cite{gaudet2021removing} noted that the Hamilton product and quaternion convolution, when used in deep networks, did not require the entire Quaternion
algebra. They called these vectormap convolutions. The weight sharing ratio is $\frac{1}{N}$ where
$N$ is the dimension of the vectormap, $D_{vm}$. Let $V_{in}^3 = [v_1,v_2,v_3]$ be an RGB input vector and $W^3 = [w_1, w_2, w_3]$ a weight vector with $N=3$. We use a permutation $\tau$ on inputs so each input vector is multiplied by each weight vector element:
\begin{equation}
\tau(v_i) = 
\begin{cases}
      v_3 & \text{$i$ = 1}\\
      v_{i-1} & \text{$i >$ 1}
    \end{cases}
\label{equ:VecinputPermutation}
\end{equation}
After applying circularly right shifted permutation to $V^3_{in}$, a new vector $V^3$ is formed. 
The permutation of weight $\tau(W^3)$ can be found like equation \ref{equ:VecinputPermutation}. 
Hence, the output vector $V_{out}$ is:
\begin{equation}
V_{out}^3 = [W^3\cdot V^3_{in}, \tau(W^3)\cdot V^3_{in}, \tau^2(W^3)\cdot V^3_{in}]
\label{equ:VecPermutedOutput}
\end{equation}
Here, ``$\cdot$'' denotes dot product. 
The outputs $V_{out}^3$ come from the linear combination of the elements of $V^3_{in}$ and $W^3$. Let the weight filter matrix for a vectormap be $V_F = [A, B, C]$ and the input vector after linear combination be
$V_h = [x, y, z]$, the vectormap convolution between $V_F$, and $V_h$ for $D_{vm} = 3$ is:
\begin{equation}
\hspace*{-2mm}
\begin{bmatrix}
 \mathscr{R}(\textbf{$V_F$}\ast \textbf{$V_h$}) \\ 
 \mathscr{I}(\textbf{$V_F$}\ast \textbf{$V_h$}) \\ 
 \mathscr{J}(\textbf{$V_F$}\ast \textbf{$V_h$}) 
\end{bmatrix}
= L \odot
\begin{bmatrix}
 \textbf{A} & \textbf{B} & \textbf{C} \\
 \textbf{C} & \textbf{A} & \textbf{B} \\
 \textbf{B} & \textbf{C} & \textbf{A} 
\end{bmatrix}
\ast
\begin{bmatrix}
 \textbf{x} \\
 \textbf{y} \\
 \textbf{z} 
\end{bmatrix}
\label{e:vectconv}
\end{equation}
where, L is a learnable matrix defined as a matrix $L \in \mathbb{R}^{D_{vm} \times D_{vm}}$ which is initialized using:
\begin{equation}
  l_{ij} =
    \begin{cases}
      1 & \text{$i$ = 1}\\
      1 & \text{$i$ = $j$}\\
      1 & \text{$j = Cal_i $ where $ Cal_i = (i+(i-1)) \And$}\\
        & \text{$Cal_i = Cal_i - D_{vm}$ if $ Cal_i > D_{vm}$} \\
      -1 & \text{else.}
    \end{cases}
\label{equ:VectConstantMatrix}
\end{equation}
By choosing $D_{vm}$ and assigning a new constant matrix $L \in \mathbb{R}^{D_{vm} \times D_{vm}}$ matching $D_{vm}$, any dimensional hypercomplex convolution can be used. 
Vectormap weight initialization uses a similar mechanism to complex \cite{trabelsi2017deep} and quaternion \cite{gaudet2018deep} weight initialization. 
Our weight initialization follows \cite{gaudet2021removing}.

\subsection{PHM Fully Connected Layer}
\label{subsection_phmconv}

The above methods apply to convolutional layers but not to fully connected (FC) layers.
\cite{zhang2021beyond} proposed parameterized
hypercomplex multiplication (PHM) for FC layers. Like vectormaps, PHM can have any dimension. 
If the dimension is four, it is like the Hamilton product. The success of the Hamiltonian product is shown in\cite{gaudet2018deep,gaudet2021removing,long2019quaternion,parcollet2018quaternion,yin2019quaternion,zhu2018quaternion}. 
Our work uses two different PHM dimensions:
four for quaternion networks, and five for vectormap networks.

A fully connected layer is defined \cite{zhang2021beyond} as $
 \textbf{y} = FC(\textbf{x}) = \textbf{W}\textbf{x} + \textbf{b}$, where
$\mathbf{W} \in \mathbb{R}^{k\times d}$ and $\mathbf{b} \in \mathbb{R}^k$ are weights and bias,
$d$ and $k$ are input and output dimensions,
and $\textbf{x}\in\mathbb{R}^d$, $\textbf{y}\in\mathbb{R}^k$.
PHM uses the following hypercomplex transform to map input $\textbf{x} \in \mathbb{R}^d$ into output $\textbf{y} \in \mathbb{R}^k$ as $
\textbf{y} = \mathit{PHM}(\textbf{x}) = \mathbf{H}\textbf{x} + \mathbf{b}$, where
$\mathbf{H} \in \mathbb{R}^{k\times d}$ is the sum of Kronecker products. 
Like $D_{vm}$, let the dimension of the PHM module be $D_{phm}=N$.
The PHM operation requires
that both $d$ and $k$ are divisible by $N$. 
$\mathbf{H}$ is the sum of Kronecker products of the parameter
matrices $\mathbf{A}_i\in\mathbb{R}^{N\times N}$
and $\mathbf{S}_i \in \mathbb{R}^{k/N\times d/N}$, where $i = 1\ldots N$:
\begin{figure*}
    \centering
    \includegraphics[width=.95\linewidth]{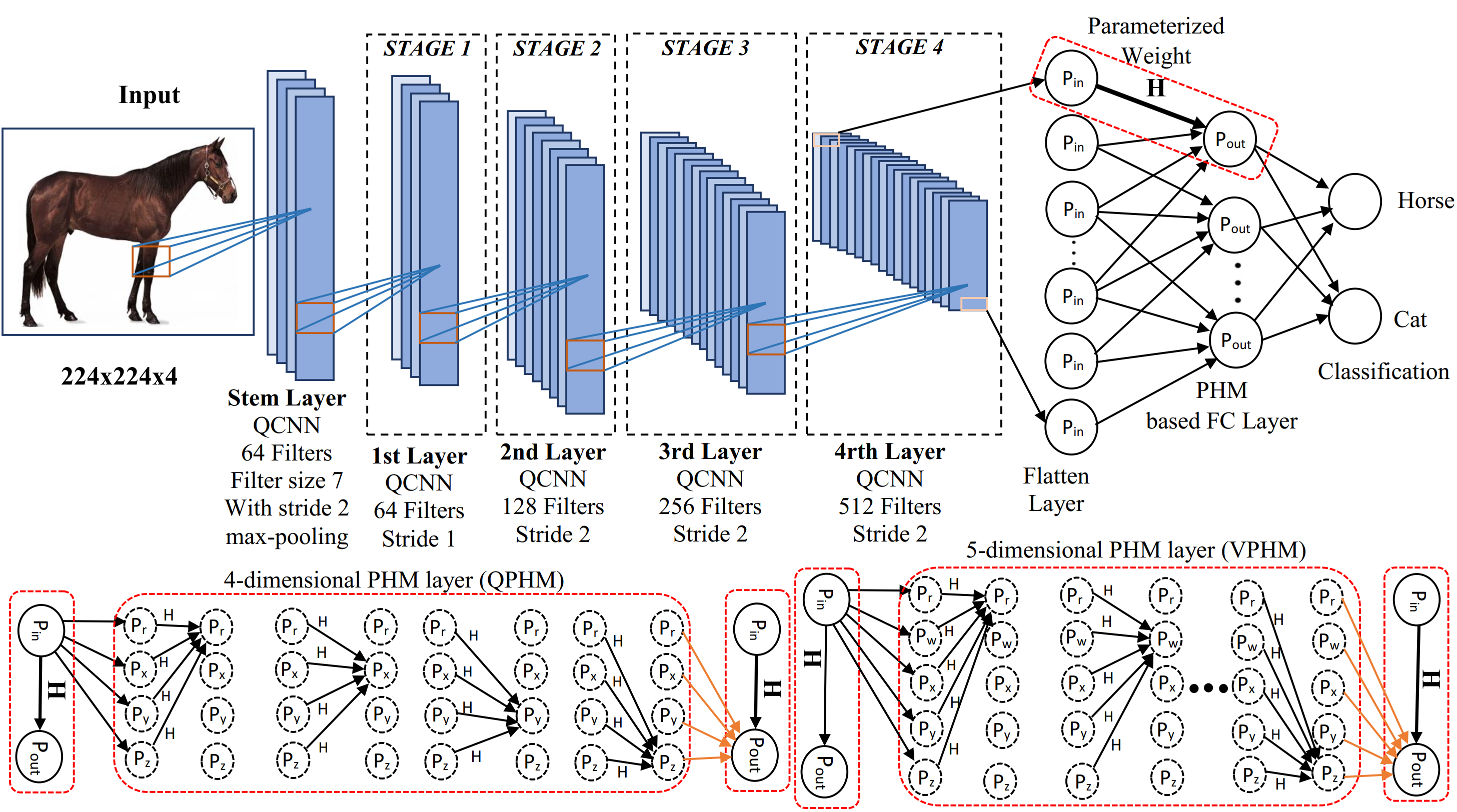}
    \caption{Full hypercomplex network where quaternion convolutional neural networks (QCNNs) are used in the front and PHM based fully-connected layers are applied in the back-end. 5-dimensional PHM is explained in Equation \ref{e:PHM5matrix}. 
    Equations \ref{equ:PHMInput} and \ref{equ:PHMOut} describe input and output for a 5D PHM layer.
    4D PHM is similar.}
    \label{fig:fig3QPHM}
\end{figure*}
\noindent
$\mathbf{H} = \sum_{i=1}^N \mathbf{A}_i \otimes \mathbf{S}_i$. 
Parameter reduction comes from reusing matrices $\mathbf{A}$ and $\mathbf{S}$ in the PHM layer.
The $\otimes$ is the Kronecker product. H is multiplied with the input in the dense layer. The four dimensional PHM layer is explained in \cite{zhang2021beyond}. We also use five dimensions which is explained here. The learnable parameters for $N=5$ are $P_r$, $P_w$, $P_x$, $P_y$, and $P_z$ where $P\in \mathbb{R}^{1\times1}$. 
For $\mathbf{A}_i$
we use the hypercomplex matrix (5 dimensions) which is generated in a similar way of vectormap convolution 
(Equations~\ref{e:vectconv} and \ref{equ:VectConstantMatrix}). 
$\mathbf{H}$ is calculated using two learnable parameter matrices ($\mathbf{A}_i$, and $\mathbf{S}_i$) for $N = 5$ as follows:
\begin{strip} 
\begin{equation}
\begin{aligned}
\mathbf{H} =
\underbrace{ 
\begin{bmatrix}\fontsize{8pt}{8pt}
 \textbf{1} & 0 & 0 & 0 & 0 \\ 0 & \textbf{1} & 0 & 0 & 0 \\  0 & 0 & \textbf{1} & 0 & 0 \\ 0 & 0 & 0 & \textbf{1} & 0 \\ 
 0 & 0 & 0 & 0 & \textbf{1}
\end{bmatrix}}_{A_1}
\otimes \underbrace{ \begin{bmatrix} P_r \end{bmatrix} }_{S_1} 
+
\underbrace{ \begin{bmatrix}
 0 & \textbf{1} & 0 & 0 & 0 \\ 0 & 0 & \textbf{1} & 0 & 0 \\ 0 & 0 & 0 & \textbf{-1} & 0 \\ 0 & 0 & 0 & 0 & \textbf{-1} \\
 \textbf{-1} & 0 & 0 & 0 & 0 
\end{bmatrix}}_{A_2}\otimes\underbrace{\begin{bmatrix} P_w\end{bmatrix}}_{S_2}
+
\underbrace{\begin{bmatrix}
 0 & 0 & \textbf{1} & 0 & 0 \\ 0 & 0 & 0 & \textbf{-1} & 0 \\ 0 & 0 & 0 & 0 & \textbf{1} \\ \textbf{-1} & 0 & 0 & 0 & 0 \\
 0 & \textbf{-1} & 0 & 0 & 0
\end{bmatrix}}_{A_3}\otimes\underbrace{\begin{bmatrix} P_x\end{bmatrix}}_{S_3}
+ \\
\underbrace{\begin{bmatrix}
 0 & 0 & 0 & \textbf{1} & 0 \\ 0 & 0 & 0 & 0 & \textbf{-1} \\ \textbf{-1} & 0 & 0 & 0 & 0 \\ 0 & \textbf{1} & 0 & 0 & 0 \\
 0 & 0 & \textbf{-1} & 0 & 0
\end{bmatrix}}_{A_4}\otimes\underbrace{\begin{bmatrix} P_y\end{bmatrix}}_{S_4}
+ 
\underbrace{\begin{bmatrix}
 0 & 0 & 0 & 0 & \textbf{1} \\ \textbf{-1} & 0 & 0 & 0 & 0 \\ 0 & \textbf{-1} & 0 & 0 & 0 \\ 0 & 0 & \textbf{-1} & 0 & 0 \\
 0 & 0 & 0 & \textbf{1} & 0
\end{bmatrix}}_{A_5}\otimes\underbrace{\begin{bmatrix} P_z\end{bmatrix}}_{S_5}
= 
\begin{bmatrix} P_r & P_w & P_x & P_y & P_z \\ -P_z & P_r & P_w & -P_x & -P_y \\ -P_y & -P_z & P_r & -P_w & P_x \\
 -P_x & -P_y & -P_z & P_r & -P_w \\ -P_w & -P_x & -P_y & P_z & P_r 
\end{bmatrix}
\label{e:PHM5matrix}
\end{aligned}
\end{equation}
\end{strip}

\noindent
Equation \ref{e:PHM5matrix} for $N = 5$ expresses the Hamiltonian product of hypercomplex layer. It preserves all PHM layer properties. 
 
\begin{table*}
\centering
\begin{tabular}{c|c|c|c|c|c} 
\hline
\shortstack{Layer} & \shortstack{Output \\ size} & 
\shortstack{Quaternion\\ResNet} & \shortstack{Vectormap\\ResNet} & \shortstack{QPHM} & \shortstack{VPHM} \\ \hline
\shortstack{Stem} & 32x32 & 3x3Q, 112, std=1 & 3x3V, 90, std=1 & 3x3Q, 112, std=1 & 3x3V, 90, std=1  \\ \hline

\shortstack{ Bottleneck\\ group 1} &$32\mathrm{x}32$& 
 $ \begin{bmatrix} 1\mathrm{x}1\mathrm{Q},112 \\ 3\mathrm{x}3\mathrm{Q},112 \\ 1\mathrm{x}1\mathrm{Q},448 \end{bmatrix}$ ${\times} 3 $ & 
$ \begin{bmatrix} 1\mathrm{x}1\mathrm{V}, 90 \\ 3\mathrm{x}3\mathrm{V}, 90 \\ 1\mathrm{x}1\mathrm{V}, 360 \end{bmatrix}$ ${\times} 3 $ & 
$\begin{bmatrix} 1\mathrm{x}1\mathrm{QP},112 \\ 3\mathrm{x}3\mathrm{QP},112 \\ 1\mathrm{x}1\mathrm{QP},448 \end{bmatrix}$ ${\times} 3 $ &
$ \begin{bmatrix} 1\mathrm{x}1\mathrm{VP},90 \\ 3\mathrm{x}3\mathrm{VP},90 \\ 1\mathrm{x}1\mathrm{VP},390 \end{bmatrix}$ ${\times} 3 $ 
\\ \hline

\shortstack{Bottleneck\\group 2} & $16\mathrm{x}16$ &  
$\begin{bmatrix} 1\mathrm{x}1\mathrm{Q}, 224 \\ 3\mathrm{x}3\mathrm{Q}, 224 \\ 1\mathrm{x}1\mathrm{Q}, 896 \end{bmatrix}$ ${\times} 4 $ &
$ \begin{bmatrix} 1\mathrm{x}1\mathrm{V}, 180 \\ 3\mathrm{x}3\mathrm{V}, 180 \\ 1\mathrm{x}1\mathrm{V}, 720 \end{bmatrix}$ ${\times} 4 $ &
$\begin{bmatrix} 1\mathrm{x}1\mathrm{QP}, 224 \\ 3\mathrm{x}3\mathrm{QP}, 224 \\ 1\mathrm{x}1\mathrm{QP}, 896 \end{bmatrix}$ ${\times} 4 $ &
$\begin{bmatrix} 1\mathrm{x}1\mathrm{VP},180 \\ 3\mathrm{x}3\mathrm{VP},180 \\ 1\mathrm{x}1\mathrm{VP},720 \end{bmatrix}$ ${\times} 4 $ \\ \hline

\shortstack{ Bottleneck\\group 3} & $8\mathrm{x}8$ &
$\begin{bmatrix} 1\mathrm{x}1\mathrm{Q}, 448 \\ 3\mathrm{x}3\mathrm{Q}, 448 \\ 1\mathrm{x}1\mathrm{Q}, 1792 \end{bmatrix}$ ${\times} 6 $& 
$ \begin{bmatrix} 1\mathrm{x}1\mathrm{V}, 360 \\ 3\mathrm{x}3\mathrm{V}, 360 \\ 1\mathrm{x}1\mathrm{V}, 1440 \end{bmatrix}$ ${\times} 6 $& 
$\begin{bmatrix} 1\mathrm{x}1\mathrm{QP}, 448 \\ 3\mathrm{x}3\mathrm{QP}, 448 \\ 1\mathrm{x}1\mathrm{QP}, 1792 \end{bmatrix} $ ${\times} 6 $ &
$\begin{bmatrix} 1\mathrm{x}1\mathrm{VP},360 \\ 3\mathrm{x}3\mathrm{VP},360 \\ 1\mathrm{x}1\mathrm{VP},1440 \end{bmatrix}$ ${\times} 6 $ 
\\ \hline

\shortstack{ Bottleneck\\group 4} & $4\mathrm{x}4$ & 
$\begin{bmatrix} 1\mathrm{x}1\mathrm{Q}, 896 \\ 3\mathrm{x}3\mathrm{Q}, 896 \\ 1\mathrm{x}1\mathrm{Q}, 3584\end{bmatrix}$ ${\times}3 $ &
$ \begin{bmatrix} 1\mathrm{x}1\mathrm{V}, 720 \\ 3\mathrm{x}3\mathrm{V}, 720 \\ 1\mathrm{x}1\mathrm{V}, 2880 \end{bmatrix}$ ${\times} 3 $ &
$\begin{bmatrix} 1\mathrm{x}1\mathrm{QP}, 896 \\ 3\mathrm{x}3\mathrm{QP}, 896 \\ 1\mathrm{x}1\mathrm{QP}, 3584 \end{bmatrix} $ ${\times} 3 $ &
$\begin{bmatrix} 1\mathrm{x}1\mathrm{VP}, 720 \\ 3\mathrm{x}3\mathrm{VP},720 \\ 1\mathrm{x}1\mathrm{VP},2880 \end{bmatrix}$ ${\times} 3 $ \\ \hline

\shortstack{Pooling \\ Layer}& $1\mathrm{x}1\mathrm{x}100$ & \multicolumn{3}{c}{global average-pool, 100 outputs}& \\ \hline

Output& $1\mathrm{x}1\mathrm{x}100$ & \multicolumn{2}{c|}{fully connected Layer, softmax}& QPHM Layer & VPHM Layer \\ \hline
\end{tabular}

\caption{The 50-layer architectures tested on CIFAR-100: quaternion ResNet \cite{gaudet2018deep,gaudet2021removing}, 
vectormap ResNet \cite{gaudet2021removing}, our proposed QPHM, and VPHM. 
Input is a 32x32x3 color image. 
The number of stacked bottleneck modules is specified by multipliers. 
``Q'', ``V'', ``QP'', ``VP'', and ``std'' denote quaternion convolution, vectormap convolution, 
QPHM (quaternion network with PHM layer), VPHM  (vectormap network with PHM layer), and stride respectively. 
Integers (e.g., 90, 112) denote number of output channels.
\label{tab_archiTable}
}
\end{table*}
\noindent

\section{Proposed Models: QPHM and VPHM}
\label{sec:QVPHM} 

We propose a new fully hypercomplex model in lieu of hypercomplex CNNs that use a real-valued backend dense layer. That is, we replace the dense layer with a PHM layer to enjoy the benefits of hypercomplex weight sharing.

We chose two base hypercomplex models for the convolutional frontend, the quaternion network and vectormap network \cite{gaudet2018deep,gaudet2021removing} which were using real-valued backend layers. 
To match dimensions with frontend networks, we used a PHM layer at four dimensions with the quaternion network and a PHM layer at five dimensions with the three dimensional vectormap network. 
In some cases, we also needed to use a PHM layer at five dimensions with quaternion networks. 
But we couldn't use a three dimensional PHM layer as the output classes must be divisible by the dimensions in the PHM operation. 

Figure \ref{fig:fig3QPHM} shows our proposed PHM based FC layer with quaternion convolutional neural networks (QCNNs). At the end of QCNNs (end of layer 4 in Figure \ref{fig:fig3QPHM} (top)), 
the output feature maps are flattened. This flattened layer is normally the input to a fully connected layer, 
but in our proposed method this layer is the input layer for the PHM based FC layer. This is represented as $P_{in}$. The parameterized weight $\mathbf{H}$ 
performs parameterized multiplication to find the hyper-complex output $P_{out}$. The type of PHM layer depends on the dimensions needed. For quaternion networks, we used dimensions four and five according to the number of classes in the datasets. The figures in Figure \ref{fig:fig3QPHM} (bottom) are expanded 4D PHM and 5D PHM layer of a single dense layer connection (red marked in Figure \ref{fig:fig3QPHM} (top)). 
\begin{equation}
    P_{in} = Pr_{in}+Pw_{in}+Px_{in}+Py_{in}+Pz_{in}
\label{equ:PHMInput}
\end{equation}
For the PHM layer with five dimensions, each PHM layer accepts five channels of input like $Pr_{in}$, $Pw_{in}$, $Px_{in}$, $Py_{in}$, and $Pz_{in}$ (Equation \ref{equ:PHMInput}) and produces five channels of output like $Pr_{out}$, $Pw_{out}$, $Px_{out}$, $Py_{out}$, and $Pz_{out}$ which are merged or stacked together to $P_{out}$ as,
\begin{equation}
    P_{out} = Pr_{out}+Pw_{out}+Px_{out}+Py_{out}+Pz_{out}
    \label{equ:PHMOut}
\end{equation}
Hence, the representational feature maps persist throughout the classification network. Similarly, this PHM (both 4D, and 5D) dense layer is applied in the backend of original ResNet \cite{he2016deep} which we named RPHM (ResNet-with-PHM). 

\begin{table*}
\centering
\begin{tabular}{|l|c|c|c|c|c|} \hline
\multirow{2}{*}{\shortstack{Model Name}} &
\multirow{2}{*}{\shortstack{Param  Count}} & \multirow{2}{*}{\shortstack{FLOPS}} & \multirow{2}{*}{\shortstack{Latency}} & \multicolumn{2}{c|}{Validation Accuracy} \\ \cline{5-6}
&&&& CIFAR-10 & CIFAR-100 \\ \hline
ResNet18 \cite{he2016deep}   & 11.1M & 0.56G & 0.22ms & 94.08 & 72.19 \\
RPHM18  & 11.1M & 0.55G & 0.21ms & 94.74 & 77.83 \\
Quat18 \cite{gaudet2018deep} & 8.5M & 0.26G & 0.36ms & 94.08 & 71.23 \\
Vect18 \cite{gaudet2021removing} & 7.3M & 0.21G & 0.29ms & 93.95 & 72.82 \\
QPHM18 & 8.5M & 0.25G & 0.35ms & \textbf{95.03} & \textbf{77.88} \\
VPHM18 & 7.3M & 0.20G & 0.27ms & 94.97 & 77.80\\\hline

ResNet34 \cite{he2016deep} & 21.2M & 1.16G & 0.29ms & 94.27 & 72.19 \\
RPHM34  & 21.1M & 1.15G & 0.28ms & 94.98 & 77.80 \\
Quat34 \cite{gaudet2018deep} & 16.3M & 0.438G & 0.57ms & 94.27 & 72.76 \\
Vect34 \cite{gaudet2021removing} & 14.04M & 0.35G & 0.45ms & 94.45 & 74.12 \\
QPHM34 & 16.3M & 0.432G & 0.54ms  & 95.40 & \textbf{78.51} \\
VPHM34 & 14.03M & 0.34G & 0.44ms & \textbf{95.41} & 77.23\\\hline

ResNet50 \cite{he2016deep} & 23.5M & 1.30G & 0.478ms & 93.90 & 72.60 \\
RPHM50 & 20.6M & 1.29G & 0.468ms & \textbf{95.59} & 79.21 \\
Quat50 \cite{gaudet2018deep} & 18.08M & 1.45G & 0.97ms & 93.90 & 72.68 \\
Vect50 \cite{gaudet2021removing} & 15.5M & 1.19G & 0.77ms & 94.28 & 74.84 \\
QPHM50 & 18.07M & 1.44G & 0.96ms  & \textbf{95.59} & \textbf{80.25} \\
VPHM50 & 15.5M & 1.15G & 0.76ms & 95.48 & 78.91\\\hline
\end{tabular}
\caption{Image classification performance on the CIFAR benchmarks for 18, 34 and 50-layer architectures. Here, Quat, Vect, 
QPHM, and VPHM, define the quaternion ResNet, vectormap ResNet, quaternion networks with PHM FC layer, and vectormap networks with PHM FC layer, respectively.}
\label{tab_resultTableBase}
\end{table*}
\noindent

\section{Experiment}
\label{sec:experiment}
The purpose of the experiments reported herein was to test whether replacing the
real-valued backend of a CNN model with a PHM backend improved classification performance.
The architectures tested were real-valued, 
quaternion-valued\cite{gaudet2018deep,parcollet2018quaternion}, and vectormap ResNet \cite{gaudet2021removing},
either with or without the PHM backend.
We refer to the quaternion ResNet model with the PHM backend as QPHM.
Similarly, VPHM, RPHM denote the vectormap ResNet, and real-valued ResNet models with the PHM backend.

Our experiments were conducted on the following datasets: 
CIFAR-10/100 \cite{krizhevsky2009learning}, 
the ImageNet300k dataset \cite{shahadat2021adding} 
and the American Sign Language Hand Gesture color image recognition dataset  \cite{dong2015american}. 
The first two datasets have less training samples with small image resolutions and the other 
datasets use a large number of training samples with higher resolution images. 
We used these datasets 
to check our proposed models for small and large training samples as well as for small and high resolution images. 
The experiments were run on a workstation 
with an Intel(R) Core(TM) i9-9820X CPU @ 3.30GHz, 128 GB memory, and NVIDIA Titan RTX GPU (24GB).

\noindent
\subsection{CIFAR Classification}
In addition to testing the PHM with real-valued, quaternion-valued, and vectormap ResNet, 
we tested the network models with three depths: 18, 34, and 50 layers.

\subsubsection{Method}
\label{cifarMethod}

We tested all of the above mentioned architectures with and without the PHM backend
on both the CIFAR-10 and CIFAR-100 datasets. 
These datasets were composed of 32x32 pixel RGB images falling into either ten classes
or 100 classes, respectively.
Both datasets have 50,000 training, and 10,000 test examples.

The models were trained using the same components as the real-valued networks, the original quaternion network, 
and the original vectormap network using the same datasets. 
All models in Table \ref{tab_resultTableBase} were trained using the same hyperparameters.  
Our QPHM and VPHM design is similar to the quaternion \cite{gaudet2018deep,gaudet2021removing}, and vectormap networks \cite{gaudet2021removing}, respectively. 
The residual architectures differ
in the number of output 
channels than the original hypercomplex networks and the proposed networks due to 
keeping the number of trainable parameters about the same. 
The number of output channels for the residual networks is the same as \cite{gaudet2021removing} and \cite{parcollet2018quaternion}.  
Table~\ref{tab_archiTable} shows the 50-layer architectures tested for CIFAR-100 dataset.

One goal is to see if the representations generated by the PHM based dense layer instead of the real-valued dense 
layer outperforms the quaternion, vectormap, and residual baselines reported in \cite{gaudet2021removing}.
We also analyzed different residual architectures to assess the effect of depth on our proposed models. 
For preprocessing, we followed \cite{gaudet2021removing}. 
We used stochastic gradient descent optimization with 0.9 
Nesterov momentum. 
The learning rate was initially set to 0.1 with warm-up learning for the first 10 epochs. 
For smooth learning, we chose cosine learning from epochs 11 to 120. 
However, we were getting about same performance for linear learning. 
All models were trained for 120 epochs and batch size was set to 100. 
This experiment used batch normalization and 0.0001 weight decay. 
The implementation is on github at-https://github.com/nazmul729/QPHM-VPHM.git.
\begin{figure*} 
\centering
\begin{subfigure}{0.48\linewidth}
  \centering
  \includegraphics[width=.95\linewidth]{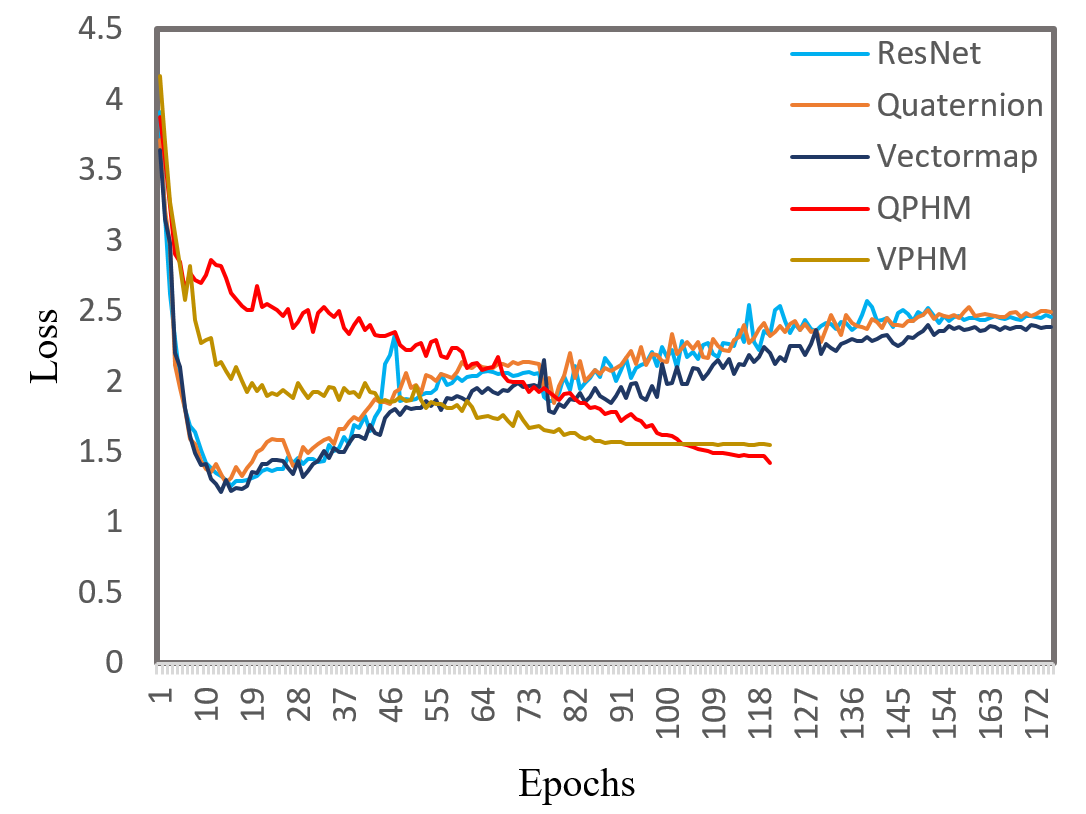}
  \captionsetup{width=0.95\linewidth}
  \captionof{figure}{Validation loss versus training.}
  \label{fig:figLoss}
\end{subfigure} 
\begin{subfigure}{0.48\linewidth}
  \centering
  \includegraphics[width=.95\linewidth]{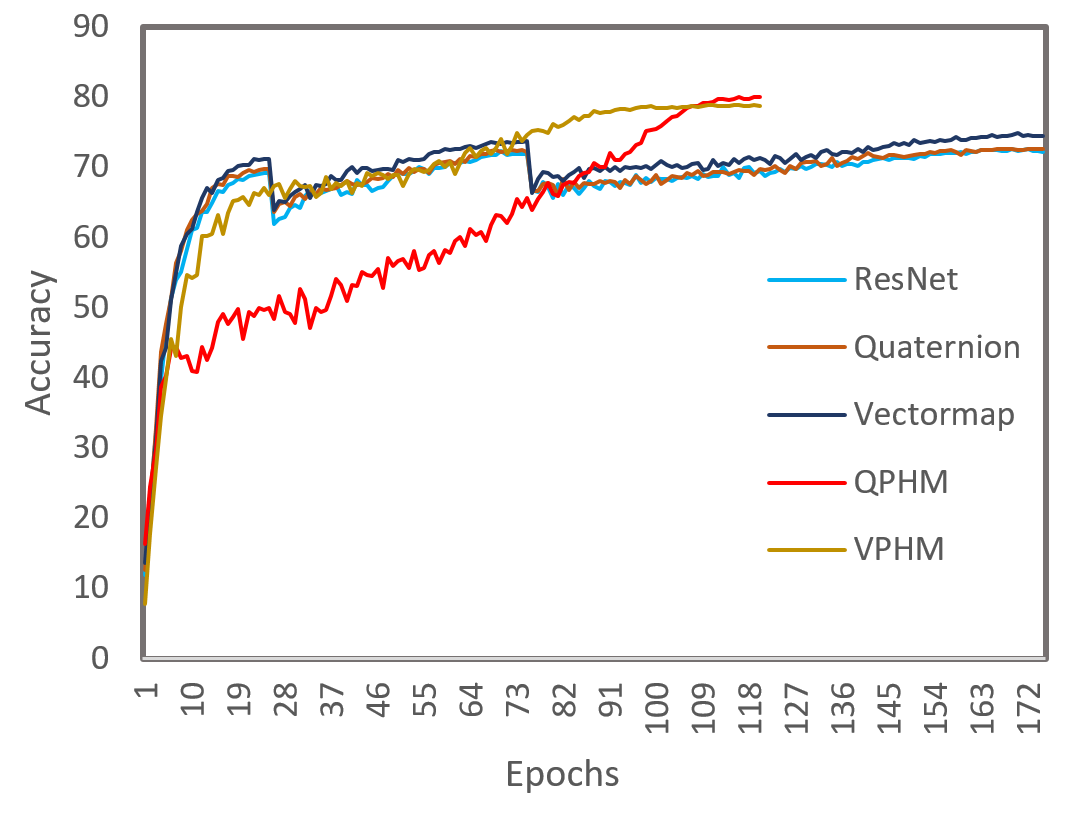}
  \captionsetup{width=0.95\linewidth}
  \captionof{figure}{Validation accuracy versus training.}
  \label{fig:figAccuracy}
\end{subfigure}
\caption{Validation loss and accuracy of 50 layer ResNet \cite{gaudet2021removing}, quaternion \cite{gaudet2021removing}, vectormap \cite{gaudet2021removing}, QPHM, VPHM for CIFAR-100.}
\label{fig_loss_and_accuracy_CIFAR100}
\end{figure*}

\subsubsection{Results}
The main results appear in Figure~\ref{fig_Fig1_CIFAR10_and100}
and~\ref{fig_loss_and_accuracy_CIFAR100}, and in Table~\ref{tab_resultTableBase}.
Figure~\ref{fig_Fig1_CIFAR10_and100} gives the overall pattern of results.
Figure~\ref{fig:fig1CIFAR10} shows results for CIFAR-10.
It shows top-1 validation accuracy for the five models:
real-valued ResNet, quaternion-valued ResNet, vectormap ResNet,
QPHM, and VPHM.
Also, results are shown for 18, 34, and 50 layers. 
We chose top-1 performance out of three.
Figure~\ref{fig:fig2CIFAR100} shows the same consistent pattern of results
for the CIFAR-100 dataset.
The magnitude of improvement is higher for CIFIR-100 than for CIFAR-10.
The results are also shown in tabular form in Table~\ref{tab_resultTableBase}, along with counts of
trainable parameters, flops, and latency.
It can be seen in Table~\ref{tab_resultTableBase} that modifying the backend to have a PHM layers has little effect
on the parameter count, flops, and latency as the input image resolutions, and the number of output classes are low.

The proposed QPHM model attains better top-1 validation accuracy than the original ResNet, quaternion, and vectormap networks for both datasets. 
The QPHM also produces better performance compared to the proposed VPHM, and RPHM models. Moreover, we compare our best performance which is obtained by the QPHM model, to the deep or shallow complex or hyper-complex networks and notice that the QPHM is achieved SOTA performance (shown in Table \ref{tab_resultCompComparison}) for the CIFAR-10 and -100 datasets. 

Table \ref{tab_resultCompComparison} compares different complex or hypercomplex networks top-1 validation accuracy with our best result. 
Our comparison was not limited to \cite{gaudet2021removing} and complex space. 
The QPHM also gains highest top-1 validation accuracy than the 
relevant CNN models for both datasets (shown in Table \ref{tab_resultRealComparison}). 
Tables \ref{tab_resultTableBase}, \ref{tab_resultCompComparison}, and \ref{tab_resultRealComparison}, 
show that our QPHM model achieves best performance for CIFAR 10 and 100 datasets with fewer parameters, flops, and latency.

\subsection{ImageNet Classification}
\subsubsection{Method}
These experiments are performed on a 300k subset of the ImageNet dataset 
which we call ImageNet300k\cite{shahadat2021adding}.
\cite{shahadat2021adding} explains how the full dataset was sampled.
The models compared are: standard ResNets \cite{shahadat2021adding}, quaternion convolutional ResNets \cite{shahadat2021adding}, and our proposed QPHM. 
We ran 26, 35, and 50-layers architectures using ``[1, 2, 4, 1]'', ``[2, 3, 4, 2]'' and ``[3, 4, 6, 3]'' bottleneck block multipliers.  
Training (all models in Table \ref{tab_resultImageNetTable}) used  the  same  optimizer  and hyperparameters as CIFAR classification method.

\begin{table}
\centering
\begin{tabular}{|l|c|c|} \hline
\multirow{2}{*}{\shortstack{Model Architecture}} & \multicolumn{2}{c|}{Validation Accuracy} \\ \cline{2-3} & CIFAR-10 & CIFAR-100 \\ \hline
DCNs \cite{anderson2017split} & 38.90 & 42.6 \\
DCN \cite{trabelsi2017deep} & 94.53 & 73.37 \\ \hline 
QCNN \cite{long2019quaternion} & 77.48 & 47.46 \\
Quat \cite{zhu2018quaternion} & 77.78 & - \\
QCNN \cite{yin2019quaternion} & 83.09 & - \\
QCNN* \cite{yin2019quaternion} & 84.15 & - \\
\hline
Quaternion18 \cite{gaudet2021removing}  & 94.80 & 71.23 \\
Quaternion34\cite{gaudet2021removing}  & 94.27 & 72.76 \\
Quaternion50\cite{gaudet2021removing}  & 93.90 & 72.68 \\\hline
Octonion \cite{wu2020deep} & 94.65 & 75.40 \\ \hline
Vectormap18\cite{gaudet2021removing}   & 93.95 & 72.82 \\
Vectormap34\cite{gaudet2021removing}    & 94.45  & 74.12 \\
Vectormap50\cite{gaudet2021removing}  & 94.28  & 74.84 \\\hline
QPHM-50 & \textbf{95.59} & \textbf{80.25}  \\
VPHM-50  & 95.48 & 78.91 \\\hline
\end{tabular}
\caption{Top-1 validation accuracy for hypercomplex networks. DCN stands for deep complex convolutional network. * variant used quaternion batch normalization. Quaternion and vectormap networks 
are the base networks
\cite{gaudet2021removing}}
\label{tab_resultCompComparison}
\end{table}

\subsubsection{Experimental Results}
Table \ref{tab_resultImageNetTable} shows the results on the ImageNet300k dataset. 
This result shows that our model takes three millions fewer trainable parameters and yields almost 5\% higher validation performance for the same architectures. Parameter reduction is not depicted in Table \ref{tab_resultCompComparison} for low resolution CIFAR benchmark images as they have saved parameters in thousands.
It is also clear that deeper networks perform better than shallow networks. 

\subsection{ASL Classification}
\subsubsection{Method}

To compare the proposed QPHM model with other networks, 
we evaluated it on the ASL Alphabet dataset \cite{upadhyay2020sign} publicly available on Kaggle at https://www.kaggle.com/grassknoted/asl-alphabet. 
This dataset has 87,000 hand-gesture images for 29 sign classes where each class has about 3,000 images. And, the image size is 200 $\times$ 200 $\times$ 3.

It has 26 finger spelling alphabet classes for the English alphabetic letters and three special characters.
Due to the divisibility restriction in PHM, we cannot use 29 classes as 29 is prime. Like all other baseline leave-one-out and half-half methods \cite{dong2015american,pugeault2011spelling,kuznetsova2013real,keskin2013real,tao2018american}, we exclude one class (letter B) from the training and validation sets. We use the same hyperparameters as CIFAR classification method.

\begin{table}  
\centering
\begin{tabular}{|l|c|c|} \hline
\multirow{2}{*}{\shortstack{Model Architecture}} & \multicolumn{2}{c|}{Validation Accuracy} \\ \cline{2-3} & CIF10 & CIF100 \\ \hline

\multicolumn{3}{|l|}{\textbf{\textit{Convolutional Networks}}} \\ \hline
ResNet18 \cite{hassani2021escaping} & 90.27 & 63.41\\
ResNet34 \cite{hassani2021escaping} & 90.51 & 64.52\\
ResNet50 \cite{hassani2021escaping} & 90.60 & 61.68\\
ResNet110 \cite{hassani2021escaping}  & 95.08 & 76.63\\ 
ResNet1001 \cite{he2016identity}  & 95.08 & 77.29 \\
MobileNet \cite{hassani2021escaping} & 91.02 & 67.44 \\ 
\hline 
Cout\cite{devries2017improved} & 95.28 & 77.54 \\\hline
\multicolumn{3}{|l|}{\textbf{\textit{Wide Residual Networks}}} \\ \hline
WRN-28-10 & 96.00 & 80.75 \\
WRN-28-10-dropout &	96.11 &	81.15 \\\hline

\multicolumn{3}{|l|}{\textbf{\textit{Our Method}}}\\\hline
QPHM50 & \textbf{95.59} & \textbf{80.25} \\
VPHM50  & 95.48 & 78.91 \\
QPHM-18-2 (ours) & \textbf{96.24} & \textbf{81.45} \\
QPHM-50-2 (ours) & \textbf{96.63} & \textbf{82.00} \\\hline
\end{tabular}
\caption{Top-1 validation accuracy comparison among deep networks. CIF10 and CIF100 stand for CIFAR10 and CIFAR100. Cout is the ResNet-18+cutout.  WRN-28-10\cite{zagoruyko2016wide}, QPHM-18-2, and QPHM-50-2 stand for wide ResNet 28, 18, and 50-layers with the output channel widening factor 10, 2, and 2, respectively.} 
\label{tab_resultRealComparison}
\end{table}

\begin{table*}
\centering
\begin{tabular}{|l|c|c|c|c|c|c|} \hline
\shortstack{Architecture} & \shortstack{Params} & \shortstack{FLOPS} & \shortstack{Latency} & \shortstack{Training\\ Accuracy} & \shortstack{Validation\\ Accuracy}\\\hline 
ResNet26 & 13.6M & 1.72G & 0.75ms & 57.0 & 45.48 \\
Quat ResNet26 & 15.1M & 1.30G & 1.71ms & 64.1 & 50.09\\
QPHM26 & \textbf{11.4M}& 1.18G & 1.7ms & \textbf{65.3} & \textbf{52.23}\\\hline
ResNet35 & 18.5M & 3.57G & 1.02ms  & 63.8 & 48.99\\
Quat ResNet35 & 20.5M & 4.59G & 3.15ms & 70.9 & 48.11\\
QPHM35 & \textbf{17.5M}& 4.10G & 3.15ms & \textbf{75.3} & \textbf{51.84} \\\hline
ResNet50 & 25.5M & 4.01G & 1.46ms & 65.8 & 50.92\\
Quat ResNet50 & 27.6M & 5.82G & 4.21ms & 73.4 & 49.69\\
QPHM50 & \textbf{24.5M} & 5.32G & 4.16ms & \textbf{78.8}  & \textbf{54.38}  \\\hline
\end{tabular}
\caption{Classification performance on the ImageNet300k dataset for different ResNet architectures. 
Top-1 training and validation accuracies.}
\label{tab_resultImageNetTable}
\end{table*}

\begin{table}
\centering
\begin{tabular}{|l|c|c|c|} \hline
Architecture  &  \shortstack{Top-1 Validation Accuracy} \\\hline
CNNs   & 82\% \\ \hline
HOG-LBP-SVM  & 98.36\% \\
HT with CNN  & 96.71\% \\ \hline
RF-JA with l-o-o   & 70\%\\
RF-JA with h-h & 90\% \\ \hline
GF-RF l-o-o  & 49\% \\
GF-RF h-h  & 75\% \\ \hline
ESF-MLRF l-o-o  & 57\% \\
ESF-MLRF h-h & 87\% \\ \hline
RF-JP l-o-o  & 43\% \\
RF-JP h-h  & 59\% \\ \hline
Faster RCNN  & 89.72\% \\
RCNNA  & 94.87\% \\ \hline
DBN  & 79\% \\
CMVA and IF l-o-o & 92.7\% \\
CMVA and IF h-h  & 99.9\% \\
CNN with ASL & 97.82\% \\ \hline
QPHM  & 100.0 \\\hline
\end{tabular}
\caption{Top-1 validation accuracy comparison with other works on ASL dataset. Here, l-o-o, h-h, HT with CNN \cite{ranga2018american,garcia2016real}, CMVA \cite{tao2018american}, RF-JA \cite{dong2015american}, GF-RF \cite{pugeault2011spelling}, ESF-MLRF \cite{kuznetsova2013real}, RF-JP \cite{keskin2013real}, RCNN \cite{upadhyay2020sign}, RCNNA \cite{upadhyay2020sign}, DBN \cite{rioux2014sign}, and HOG-LBP-SVM \cite{nguyen2019deep} mean Leave one out, half-half, HYBRID TRANSFORM, CNNs \cite{ameen2017convolutional} with multiview augmentation and IF Inference Fusion, Random Forest with Joint Angles, Gabor Filter-based features with Random Forest, Ensemble of Shape Function with Multi-Layer Random Forest, Random Forest with Joint Positions, Recurrent convolutional neural networks, Recurrent convolutional neural networks with attention, Deep belief network, and Histogram of Oriented Gradients (HOG) and Local Binary Pattern (LBP) with support vector machine, respectively.}
\label{tab_resultASLTable}
\end{table}

\subsubsection{Experimental Results}
Due to the divisibility limitation, it is not possible to evaluate the ASL data using VPHM model as we choose PHM with five dimensions for VPHM model. 
So we only tested the QPHM (PHM with four dimensions) model to compare with other networks on the ASL dataset. 
Table \ref{tab_resultASLTable} provides a comparison of top-1 validation accuracy of our proposed QPHM model with other networks in ASL data. 
Our proposed architecture performs state-of-the-art accuracy in this ASL dataset. 
Hence, the representation feature maps in the dense layer are very effective for this dataset.

\section{Conclusions}
We replaced the dense backend of existing hypercomplex CNNs for image classification with PHM modules to 
create weight sharing in this layer.
This novel design improved classification accuracy, reduced parameter counts, flops, and latency compared
to the baseline networks. The results support our hypothesis that the PHM operation in the 
densely connected back end provides better representations as well as improves accuracy with fewer parameters. These results also highlighted the importance of the calculations in the backend.

The QPHM and VPHM outperformed the other works mentioned in ``Experiment'' section. The proposed QPHM achieved higher validation accuracy (top-1) for all network architectures than the proposed VPHM.

{\small
\bibliographystyle{ieee_fullname}
\bibliography{egbib}

\begin{thebibliography}{10}\itemsep=-1pt

\bibitem{ameen2017convolutional}
Salem Ameen and Sunil Vadera.
\newblock A convolutional neural network to classify american sign language
  fingerspelling from depth and colour images.
\newblock {\em Expert Systems}, 34(3):e12197, 2017.

\bibitem{anderson2017split}
Timothy Anderson.
\newblock Split-complex convolutional neural networks.
\newblock 2017.

\bibitem{buyssens2012multiscale}
Pierre Buyssens, Abderrahim Elmoataz, and Olivier L{\'e}zoray.
\newblock Multiscale convolutional neural networks for vision--based
  classification of cells.
\newblock In {\em Asian Conference on Computer Vision}, pages 342--352.
  Springer, 2012.

\bibitem{devries2017improved}
Terrance DeVries and Graham~W Taylor.
\newblock Improved regularization of convolutional neural networks with cutout.
\newblock {\em arXiv preprint arXiv:1708.04552}, 2017.

\bibitem{dong2015american}
Cao Dong, Ming~C Leu, and Zhaozheng Yin.
\newblock American sign language alphabet recognition using microsoft kinect.
\newblock In {\em Proceedings of the IEEE conference on computer vision and
  pattern recognition workshops}, pages 44--52, 2015.

\bibitem{garcia2016real}
Brandon Garcia and Sigberto~Alarcon Viesca.
\newblock Real-time american sign language recognition with convolutional
  neural networks.
\newblock {\em Convolutional Neural Networks for Visual Recognition},
  2:225--232, 2016.

\bibitem{gaudet2021removing}
Chase Gaudet and Anthony~S. Maida.
\newblock Removing dimensional restrictions on complex/hyper-complex networks.
\newblock In {\em 2021 IEEE International Conference on Image Processing
  (ICIP)}, 2021.

\bibitem{gaudet2018deep}
Chase~J Gaudet and Anthony~S Maida.
\newblock Deep quaternion networks.
\newblock In {\em 2018 International Joint Conference on Neural Networks
  (IJCNN)}, pages 1--8. IEEE, 2018.

\bibitem{hassani2021escaping}
Ali Hassani, Steven Walton, Nikhil Shah, Abulikemu Abuduweili, Jiachen Li, and
  Humphrey Shi.
\newblock Escaping the big data paradigm with compact transformers.
\newblock {\em arXiv preprint arXiv:2104.05704}, 2021.

\bibitem{he2016deep}
Kaiming He, Xiangyu Zhang, Shaoqing Ren, and Jian Sun.
\newblock Deep residual learning for image recognition.
\newblock In {\em Proceedings of the IEEE conference on computer vision and
  pattern recognition}, pages 770--778, 2016.

\bibitem{he2016identity}
Kaiming He, Xiangyu Zhang, Shaoqing Ren, and Jian Sun.
\newblock Identity mappings in deep residual networks.
\newblock In {\em European conference on computer vision}, pages 630--645.
  Springer, 2016.

\bibitem{javanmardi2021computer}
Shima Javanmardi, Seyed-Hassan~Miraei Ashtiani, Fons~J Verbeek, and Alex
  Martynenko.
\newblock Computer-vision classification of corn seed varieties using deep
  convolutional neural network.
\newblock {\em Journal of Stored Products Research}, 92:101800, 2021.

\bibitem{keskin2013real}
Cem Keskin, Furkan K{\i}ra{\c{c}}, Yunus~Emre Kara, and Lale Akarun.
\newblock Real time hand pose estimation using depth sensors.
\newblock In {\em Consumer depth cameras for computer vision}, pages 119--137.
  Springer, 2013.

\bibitem{krizhevsky2009learning}
Alex Krizhevsky, Geoffrey Hinton, et~al.
\newblock Learning multiple layers of features from tiny images.
\newblock 2009.

\bibitem{kuznetsova2013real}
Alina Kuznetsova, Laura Leal-Taix{\'e}, and Bodo Rosenhahn.
\newblock Real-time sign language recognition using a consumer depth camera.
\newblock In {\em Proceedings of the IEEE international conference on computer
  vision workshops}, pages 83--90, 2013.

\bibitem{long2019quaternion}
Cameron~E Long.
\newblock {\em Quaternion Temporal Convolutional Neural Networks}.
\newblock PhD thesis, University of Dayton, 2019.

\bibitem{nguyen2019deep}
Huy~BD Nguyen and Hung~Ngoc Do.
\newblock Deep learning for american sign language fingerspelling recognition
  system.
\newblock In {\em 2019 26th International Conference on Telecommunications
  (ICT)}, pages 314--318. IEEE, 2019.

\bibitem{parcollet2019quaternion}
T. Parcollet, M. Morchid, and G. Linar\'{e}s.
\newblock Quaternion convolutional networks for heterogeneous image processing.
\newblock In {\em IEEE Intl. Conf. on Acoustics, Speech and Signal Processing
  (ICASSP)}, pages 8514--8518, 2019.

\bibitem{parcollet2018quaternion}
Titouan Parcollet, Ying Zhang, Mohamed Morchid, Chiheb Trabelsi, Georges
  Linar{\`e}s, Renato De~Mori, and Yoshua Bengio.
\newblock Quaternion convolutional neural networks for end-to-end automatic
  speech recognition.
\newblock {\em arXiv preprint arXiv:1806.07789}, 2018.

\bibitem{pugeault2011spelling}
Nicolas Pugeault and Richard Bowden.
\newblock Spelling it out: Real-time asl fingerspelling recognition.
\newblock In {\em 2011 IEEE International conference on computer vision
  workshops (ICCV workshops)}, pages 1114--1119. IEEE, 2011.

\bibitem{ranga2018american}
Virender Ranga, Nikita Yadav, and Pulkit Garg.
\newblock American sign language fingerspelling using hybrid discrete wavelet
  transform-gabor filter and convolutional neural network.
\newblock {\em Journal of Engineering Science and Technology},
  13(9):2655--2669, 2018.

\bibitem{rioux2014sign}
Lucas Rioux-Maldague and Philippe Giguere.
\newblock Sign language fingerspelling classification from depth and color
  images using a deep belief network.
\newblock In {\em 2014 Canadian Conference on Computer and Robot Vision}, pages
  92--97. IEEE, 2014.

\bibitem{shahadat2021adding}
Nazmul Shahadat and Anthony~S Maida.
\newblock Adding quaternion representations to attention networks for
  classification.
\newblock {\em arXiv preprint arXiv:2110.01185}, 2021.

\bibitem{tao2018american}
Wenjin Tao, Ming~C Leu, and Zhaozheng Yin.
\newblock American sign language alphabet recognition using convolutional
  neural networks with multiview augmentation and inference fusion.
\newblock {\em Engineering Applications of Artificial Intelligence},
  76:202--213, 2018.

\bibitem{trabelsi2017deep}
Chiheb Trabelsi, Olexa Bilaniuk, Ying Zhang, Dmitriy Serdyuk, Sandeep
  Subramanian, Joao~Felipe Santos, Soroush Mehri, Negar Rostamzadeh, Yoshua
  Bengio, and Christopher~J Pal.
\newblock Deep complex networks.
\newblock {\em arXiv preprint arXiv:1705.09792}, 2017.

\bibitem{upadhyay2020sign}
Shweta Upadhyay, RK Sharma, and Prashant~Singh Rana.
\newblock Sign language recognition with visual attention.
\newblock Technical report, EasyChair, 2020.

\bibitem{wu2020deep}
Jiasong Wu, Ling Xu, Fuzhi Wu, Youyong Kong, Lotfi Senhadji, and Huazhong Shu.
\newblock Deep octonion networks.
\newblock {\em Neurocomputing}, 397:179--191, 2020.

\bibitem{yin2019quaternion}
Qilin Yin, Jinwei Wang, Xiangyang Luo, Jiangtao Zhai, Sunil~Kr Jha, and
  Yun-Qing Shi.
\newblock Quaternion convolutional neural network for color image
  classification and forensics.
\newblock {\em IEEE Access}, 7:20293--20301, 2019.

\bibitem{zagoruyko2016wide}
Sergey Zagoruyko and Nikos Komodakis.
\newblock Wide residual networks.
\newblock {\em arXiv preprint arXiv:1605.07146}, 2016.

\bibitem{zhang2021beyond}
Aston Zhang, Yi Tay, Shuai Zhang, Alvin Chan, Anh~Tuan Luu, Siu~Cheung Hui, and
  Jie Fu.
\newblock Beyond fully-connected layers with quaternions: Parameterization of
  hypercomplex multiplications with $1/n $ parameters.
\newblock {\em arXiv preprint arXiv:2102.08597}, 2021.

\bibitem{zhu2018quaternion}
Xuanyu Zhu, Yi Xu, Hongteng Xu, and Changjian Chen.
\newblock Quaternion convolutional neural networks.
\newblock In {\em Proceedings of the European Conference on Computer Vision
  (ECCV)}, pages 631--647, 2018.

\end{thebibliography}
}

\end{document}